\documentclass{article}
\usepackage{spconf,amsmath,graphicx}

\usepackage{enumitem}
\setlist{nosep, leftmargin=14pt}

\usepackage{mwe} 

\usepackage{url} 

\usepackage[dvipsnames]{xcolor}

\title{Segmentation confidence for arbitrary CNNs}
\name{Author(s) Name(s)}
\address{Author Affiliation(s)}

\name{\begin{tabular}{c}Baris Oguz$^{\star}$, Xing Yao$^{\dagger}$, Claudia Tawil$^{\star}$,
  Alison Pouch$^{\star}$, Brett Byram$^{\dagger}$ \\
  Gabriel Arenas$^{\star}$, Nadav Schwartz$^{\star}$, Ipek Oguz$^{\dagger}$\end{tabular}}

\address{$^{\star}$ University of Pennsylvania 
$^{\dagger}$ Vanderbilt University}

\begin{document}
%
\maketitle
\begin{abstract}
Convolutional neural networks (CNN) are widely used for medical image segmentation. However, they lack an inherent measure of confidence. Existing methods for estimating prediction uncertainty often require a custom network architecture and/or dedicated training setups and need to be trained from scratch. In addition, the deduction of a single confidence score per input from the uncertainty maps is not always straightforward. In this study, we explore voxel-based uncertainty estimation methods that can be added to existing segmentation pipelines without training from scratch. We also propose a novel method to estimate a single confidence score from the uncertainty maps. Our evaluations on different ROIs and modalities suggest that our confidence score can predict segmentation performance in unseen data.
\end{abstract}

\section{Introduction}

Deep learning models, and in particular CNNs,
are widely used in medical image segmentation \cite{long2015fully,ronneberger2015u,he2016deep,huang2017densely,chollet2017xception,oktay2018attention,isensee2021nnu}.
However, these models often provide little insight into their  decision process \cite{hasani2022trustworthy,chen2022explainable,jin2023guidelines}, and they lack an inherent measure of confidence.
The produced probability maps provide an indirect signal of confidence per voxel. However, there is no obvious way to produce a single measure of confidence for the entire image. Since CNNs are highly successful at segmentation, the probabilities are usually higher inside and around the ROI, and mostly 0 further from the ROI. This results in a sharp transition between the foreground and background, rather than a gradual increase, limiting the ability to use volume-wide statistics as a measure of confidence. While classification models output  probabilities per class, which can estimate confidence, this is not available for segmentation.

There has been research to overcome this shortcoming \cite{czolbe2021segmentation}, often with Bayesian frameworks.
Gal and Ghahramani \cite{gal2016dropout} have shown that CNNs trained with dropout can approximate Bayesian inference by keeping dropout active during test-time and performing multiple forward passes for classification. Ghoshal et al.\ \cite{ghoshal2021estimating} propose Monte Carlo drop weights, and decompose predictive probabilities into aleatoric and epistemic uncertainties. 
Teye et al.\  \cite{teye2018bayesian} use batch normalization layers instead of dropout. Wang et al.\ \cite{wang2019aleatoric} use test-time augmentation and Monte Carlo simulation.
There have also been non-Bayesian approaches. Some methods  \cite{lakshminarayanan2017simple,karimi2019accurate}
train an ensemble of models to estimate classification uncertainty.  Kohl et al.\  \cite{kohl2018probabilistic} propose a probabilistic U-Net. Yao et al.\  \cite{yao2023false} combine an aleatoric uncertainty-based correction strategy with semi-automated prompt augmentation.
Most of these approaches require dedicated architectures \cite{akrami2021quantile} or training an ensemble of models \cite{mehrtash2020confidence}  to accomplish segmentation and confidence/uncertainty estimation simultaneously. Many of them produce an uncertainty map, but do not deduce a single score for the entire image. This limits their usability in typical medical image analysis workflows. Notably, Wang et al.\ \cite{wang2019aleatoric} use the volume variation coefficient (VVC) of multiple forward passes to estimate structure-wise uncertainty.

While CNNs perform well overall, a poor prediction for a particular image may lead to mistakes. Flagging uncertain predictions could provide alternatives, such as obtaining a repeat scan if there is an image quality issue. Another use case is flagging uncertain outputs in large-scale studies for visual quality control, manual correction, or exclusion from downstream analysis. Confidence scores can also be useful in active learning \cite{liu2023colossal,nath2020diminishing} for prioritizing low-confidence images for annotation, as these are likely not well represented in the  training dataset.
Finally, a model's ability to report its own confidence is a significant step towards explainable AI.

In this study, we use two alternatives (Sec.\ \ref{sec:testtime}) to produce multiple predictions from a single input image, which can be readily implemented as an add-on to existing segmentation pipelines. We then explore two approaches to estimate a voxel-based uncertainty map (Sec.\ \ref{sec:uncertainty}), and propose a novel method to estimate a single, image-wide confidence score (Sec.\ \ref{sec:confidence}). We evaluate our method on placenta  (ultrasound) and hippocampus  (MRI) datasets to test its generalizability.

\section{Methods}

\subsection{Data and Segmentation Models}
\label{sec:segmentation}

\textbf{\underline{Placenta 3D ultrasound (3DUS) dataset:}} This dataset contains $N=210$ images from 121 unique subjects in the first trimester, and $N=49$ images from 41 unique subjects taken during the second trimester. Placentas were manually segmented using ITK-SNAP  (\url{www.itksnap.org}).
Images were down-sampled to 128$\times$128$\times$128 voxels before training.

Automated quantification of the placenta from 3DUS data is challenging \cite{schwartz2022fully}, and quick identification of sub-optimal segmentations during clinical screening can allow the option of a repeat scan while the patient is still in the clinic.

\noindent \textbf{\underline{Hippocampus MRI dataset:}}
We used the hippocampus task from the Medical Segmentation Decathlon  \cite{simpson2019large}. We used the $N=260$ training set with anterior and posterior hippocampus labels.
Images were up-sampled to 64$^3$ voxels.

\noindent \textbf{\underline{Segmentation:}}
We employed typical 2D and 3D U-Net models with 2 levels of down / up-sampling, and skip-connections. All convolutional layers have $3^2$ ($3^3$) pixel (voxel) kernels. The placenta dataset has a single ROI. The hippocampus dataset has 2 ROIs (anterior/posterior), and we trained a separate model per ROI. We used binary cross-entropy loss.

For our 2D models for both the placenta and the hippocampus, we observed that a single model can be successfully trained on sagittal and coronal slices, since the general shape and position of the ROI are roughly similar along these axes. Slices along the axial plane differ considerably from the other 2 planes, and  inclusion in training significantly lowered  performance. Therefore, our final 2D models were trained on sagittal and coronal slices.
The placenta model has 364K trainable parameters, while the two 2D hippocampus models each have 162K. We also trained 3D models for each hippocampus ROI (1,938K trainable parameters each).
We used 4-fold cross-validation.
Since our focus is not on segmentation accuracy, but rather on the performance of confidence scoring on the test set, we used 1 partition for training and 3 for testing within each fold to provide more test cases, rather than the usual 3:1 train/test split. The random partitioning for placentas was based on the unique subjects to avoid leakage.

\subsection{Test-time augmentation strategies}
\label{sec:testtime}

\noindent \textbf{\underline{Test-time flip augmentations:}} We used test-time augmentation with horizontal/vertical flips for the 2D models. We thus applied a combination of 4 augmentations per 2D slice. As our models were trained on both the sagittal and coronal planes, we produced 8 full-volume 3D outputs per input. 

This approach was viable for our 2D models, since the random online augmentations  \textit{during training} included the horizontal/vertical flips. Specifically, for training the placenta model, we used horizontal/vertical flips, in-plane rotations, cropping, blurring, and gamma adjustment. For the hippocampus models, we only applied horizontal/vertical flips. It is common practice to utilize random augmentations like these during training, and modifying the test-time loop to include the flip augmentations and produce multiple outputs per input is a straightforward approach to apply our confidence scoring method to existing segmentation pipelines.

\noindent \textbf{\underline{Test-time noise augmentations:}} Some models may perform worse with test-time flips, e.g., if augmentations were not used during training, or if the ROI is highly asymmetric. To test our method's applicability in such cases, we trained two 3D models for the two hippocampus ROIs without any augmentations. At test-time, we added random Gaussian noise ($torch.randn\times0.5$) to a random 20\% subset of the voxels. We blurred ($\sigma=1$)  the noise before adding to the original image for a more realistic look. For each input, we produced 15 noisy versions, resulting in 16  outputs. While other noise distributions may be more suitable for various modalities, this general approach is applicable to any existing 2D or 3D models,  whether augmentations were used during training or not.

\begin{figure}[b]
\includegraphics[width=8.5cm]{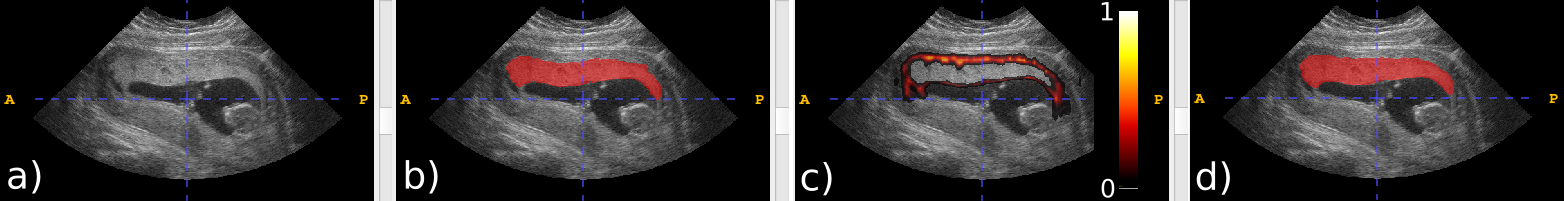}
\caption{
\textbf{a)} 3DUS placenta input, \textbf{b)} ground truth, \textbf{c)} uncertainty map $u(x)$, \textbf{d)} thresholded segmentation mask $M_0(x)$.} \label{fig4}
\end{figure}

\begin{figure}[t]
\includegraphics[width=8.5cm]{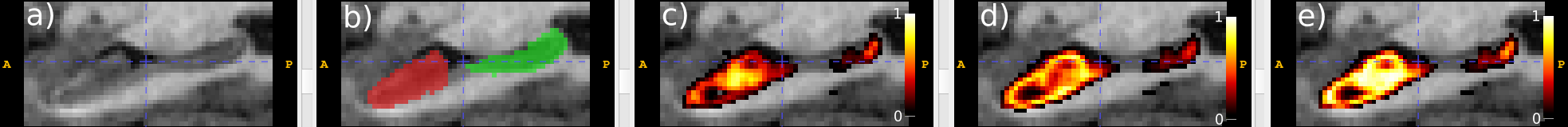}
\caption{A hippocampus sample.
\textbf{a)} 3D MRI input, \textbf{b)} ground truth: anterior (red) and posterior (green) ROIs, \textbf{c)} stdev uncertainty map, \textbf{d)} entropy uncertainty map, \textbf{e)} stdev+entropy uncertainty map. \textbf{(c-e)} show $u(x)$ for anterior ROI only.} \label{fig3hipo}
\end{figure}

\subsection{Uncertainty estimation strategies}
\label{sec:uncertainty}

\noindent \textbf{\underline{Standard deviation (stdev):}} First, we simply calculate the voxel-wise stdev of the multiple probability maps. We normalize to  [0, 1] via min-max scaling. This gives us an approximate uncertainty map $u(x)$, in which values close to 0.0 denote voxels where the multiple predictions mostly agree, and values close to 1.0 denote the most disagreement. As expected, we usually observed highest uncertainty along the boundary of the ROI (Fig.~\ref{fig4}c), as well as in  regions with artifacts  or  other noise. We used this approach in our 2D models.

\noindent \textbf{\underline{Entropy:}}
When the multiple predictions all have a probability close to 0.5 for a given voxel, since there is agreement, the low stdev registers as a low uncertainty voxel, even though the 0.5 probabilities suggest high uncertainty. To remedy this, we also calculated the binary entropy $H(p) = -p \ln p - (1 - p) \ln (1 - p)$ for all the outputs and took their voxel-wise mean. In our 3D models, we min-max scaled this mean entropy map, and averaged it with the stdev map $u(x)$ (Fig.~\ref{fig3hipo}).

\subsection{Confidence score estimation}
\label{sec:confidence}

We use the voxel-wise mean of the multiple output probability maps to create a final probability map $p(x)$ and threshold it to create a binary mask  $M_0(x)=\{p(x)\geq0.50\}$ (Fig.\ \ref{fig4}d).

\begin{figure}[b]
\includegraphics[width=8.5cm]{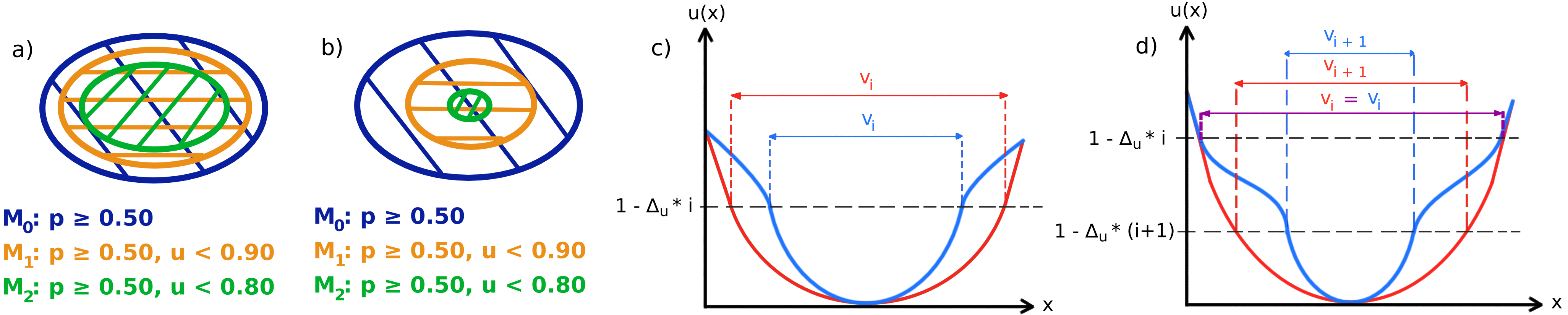}
\caption{\textbf{a)} Segmentations with low uncertainty/high confidence. \textbf{b)} Segmentations  with high uncertainty/low confidence. \textbf{c)} 1D illustration of $V_i$. The blue curve is overall more uncertain than the red curve, and thus has a lower $V_i$. \textbf{d)} Illustration of $V_{i+1}/V_i$. The blue curve has a (locally) slower change in uncertainty, and thus a smaller $V_{i+1}/V_i$.} \label{figovals} \label{uncert}
\end{figure}

Next, we calculated a base mask volume $V_0=|M_0(x)|$.
This volume is independent of the uncertainty $u(x)$, since it includes all voxels that have $p(x)\geq0.50$ regardless of uncertainty. If we introduce an uncertainty constraint such that only voxels under a given uncertainty are allowed, e.g., $u(x)\leq 0.9$, then the segmentation would be more conservative and thus the resultant mask volume would be lower than $V_0$.
We observe that if the overall model prediction for a given image $I(x)$ was highly confident, there would be very few voxels with a high uncertainty $u(x)$. Conversely, in a low-confidence prediction, there would be high uncertainty in substantially more voxels. This leads to the intuition that if the mask volume decreases rapidly as we tighten the uncertainty constraint, then the confidence for the prediction as a whole is likely low. In contrast, a very gradual decrease in volume suggests low uncertainty and high confidence (Fig.~\ref{figovals}a,~\ref{figovals}b).

To quantify this intuitive concept, we  introduced $N_u$ uncertainty constraints decreased by uniform steps $\Delta_u = 1 / N_u$, and calculated $N_u-1$ more masks similarly to $M_0$. In our experiments, we empirically used $N_u=10$ and therefore $\Delta_u=0.1$. Thus, for example, $M_1$ is the set of voxels in the segmentation that have a corresponding uncertainty of less than 0.90 ($p(x) \geq 0.50 \land u(x) \leq 0.90$), and $M_2$ only includes voxels in the mask that have an uncertainty less than 0.80. In general, for $i \in [0, N_u-1]$, we have $M_i(x)=\{p(x)\geq0.5\ \land  u(x) \leq (1 - i * \Delta_u)\}$, with volume $V_i = | M_i(x) |$.

As mentioned above, $M_0$ is the largest mask  and the volumes decrease through $V_0$ to $V_{N_u-1}$.
\textit{How} these volumes change  is informative (Fig.~\ref{figovals}a vs.\  \ref{figovals}b).
To capture this, we calculated the $N_u-1$ ratios of each volume to the base volume, $V_i/V_0$, for $i\in[1, N_u-1]$.  We also calculated the $N_u-2$ ratios of consecutive steps, $V_{i+1}/V_i$, for $i\in[1, N_u-2]$. Since the volumes are monotonically decreasing from $V_0$ through $V_{N_u-1}$, all of these ratios are in the [0, 1] range.

$V_i$ is a measure of overall certainty (Fig.~\ref{uncert}c). Dividing by $V_0$  normalizes for object size. We note that simply averaging $u(x)$ across $V_0$ would be sensitive to object size, without a straightforward way to compare different subjects. Our formulation transforms the uncertainty map $u(x)$ to a series of volumetric measures $V_i$, allowing us to compare different subjects by simply expressing the volumes as the ratio $V_i/V_0$. The $\Delta_u$ determines the granularity of this transformation, and we found 0.1 to be adequate for our purposes.
In contrast, $V_{i+1}/V_{i}$ is a measure of speed of uncertainty change (Fig.~\ref{uncert}d): the blue curve locally has slower change than the red curve, thus a smaller $V_{i+1}/V_{i}$.
To obtain a single confidence score $C$ from these ratios, we  simply took their mean, as a balanced and robust indicator: 
\begin{align}
    C = \frac{\sum_{i=1}^{N_u-1} (\frac{V_i}{V_0}) + \sum _{i=1}^{N_u-2} (\frac{V_{i+1}}{V_i}) }{(N_u-1)+(N_u-2)}
\end{align}

\subsection{Correlation with Dice and outlier detection}
\label{sec:correlation}

We examined the correlation of our confidence $C$ with the segmentation Dice.
We also used these scores to identify poor segmentations. To do this in unsupervised manner, we applied an elliptic envelope model to the confidence scores (without taking Dice as input), fitting an ellipsoid around the central cluster of the data points, to  separate inliers from outliers. We used the scikit-learn implementation of the model, with a contamination value of 0.1.
Since we were interested in only the abnormally low confidence cases, we filtered out cases with above mean confidence (i.e., the highest confidence scores) from the outlier set. 
We then performed a two-tailed t-test to assess Dice differences between inliers and outliers.

\subsection{Using confidence to select annotation candidates}
\label{sec:confselection}

We also explored how confidence scores can be used to select  images to annotate and add to the training set from a pool of unlabeled candidates. We trained a base 2D model (Sec.\ \ref{sec:segmentation}) on 21 random images from the $N=121$  placenta dataset. We split the remaining images into $N=60$ test set, and $N=40$ annotation candidates. 
Using the base model, we produced segmentations and confidence scores for the $N=40$ candidates. We compared 5 strategies for selecting candidates: best Dice, worst Dice, best confidence, worst confidence, and random. We note that the best/worst Dice strategies were included as oracle baselines, and would not be available in practice, while confidence can be computed with the base model. We added the images selected by each strategy to the $N=21$ base set in increments of 5 images, and trained new models. Since the number of images in the training set were limited, we trained 3 models for each scenario with different random seeds. We compared  performance on the $N=60$ test set.

\begin{figure}[t]

\includegraphics[width=8.5cm]{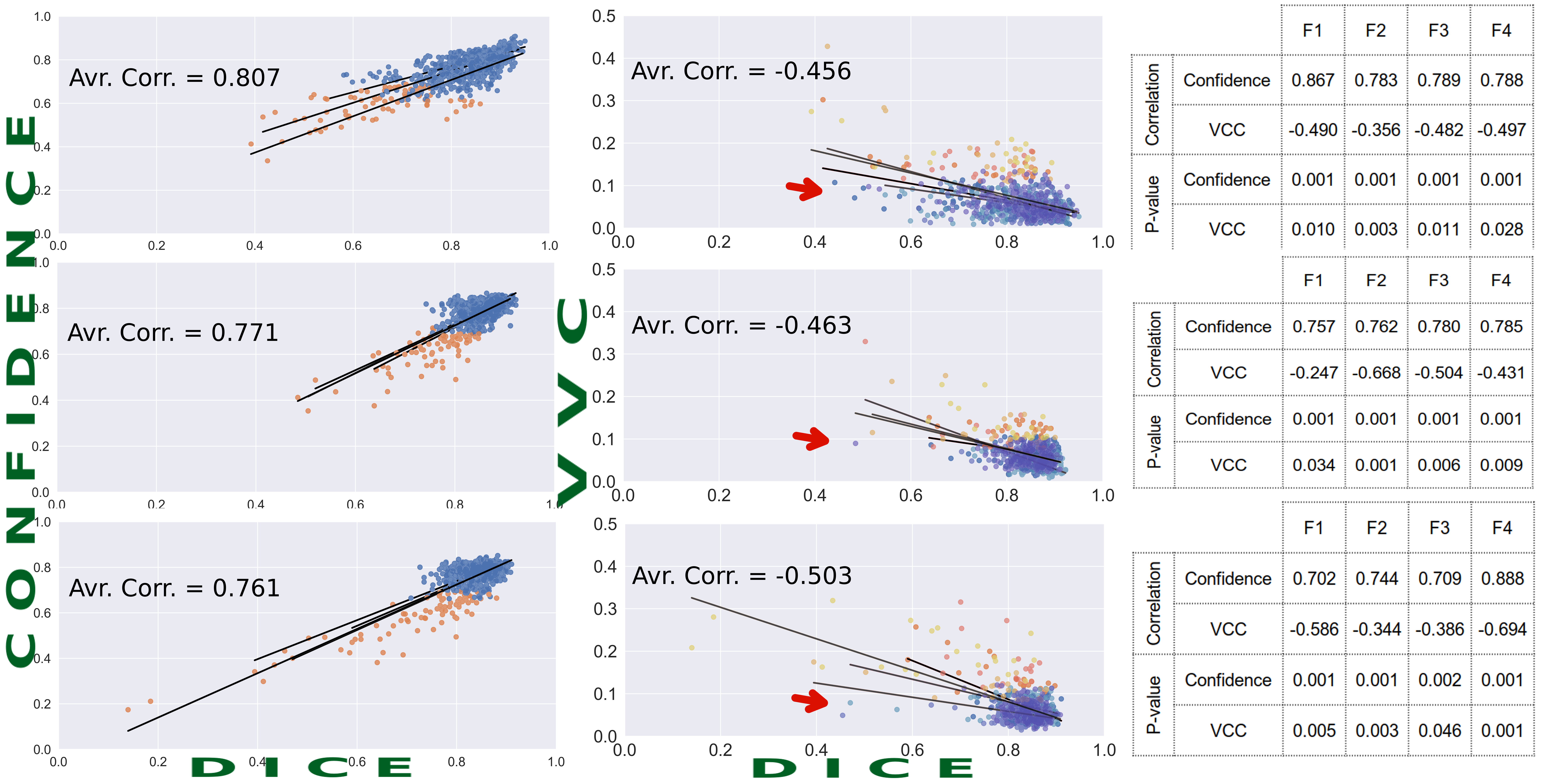}
\caption{Dice-confidence and Dice-VVC correlations. \textbf{Top,} placenta; \textbf{middle,} anterior hippocampus; \textbf{bottom,} posterior hippocampus. 2D models with flip aug., stdev uncertainty. Blue/red dots denote inliers/outliers (Sec.\ \ref{sec:correlation}), with shades corresponding to individual folds. Confidence has stronger correlation with Dice than VVC across each ROI/fold.} \label{fig-2d}
\end{figure}

\section{Results and Discussion}

\noindent \textbf{\underline{Compared methods:}} We compare our  confidence score to the volume variation coefficient (VVC) method  \cite{wang2019aleatoric} for estimating a volume-wide confidence. Other methods noted in the Introduction do not produce a volume-wide score.

\noindent \textbf{\underline{2D models:}} 
For the 2D models, we used flip augmentations during test-time, and estimated uncertainty with standard deviation (Fig.~\ref{fig-2d}). For the \underline{placenta}, the mean Dice  was between 0.81-0.82 across the 4  folds. Using confidence $C$, the mean Dice difference between inliers-outliers were in the range 0.15-0.24. We note that the outlier detection only uses $C$, and the t-test is between the associated Dice scores rather than $C$ (which would be lower in the outlier group, by construction). The inlier-outlier Dice difference was 0.07-0.10 with VVC. For \underline{anterior hippocampus}, mean Dice was between 0.84-0.85. Inlier-outlier Dice difference was 0.10-0.11 for $C$ and 0.04-0.08 for VVC. For \underline{posterior} \underline{hippocampus}, the mean Dice was between 0.82-0.84. Inlier-outlier Dice difference was 0.08-0.22 for $C$ and 0.05-0.21 for VVC.

\begin{figure}[b]
\includegraphics[width=8.5cm]{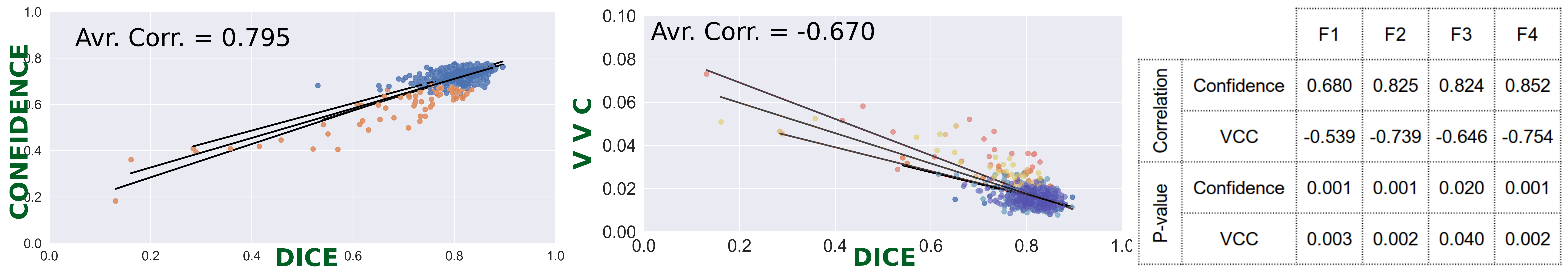}
\caption{Posterior hippocampus Dice, confidence, VVC. 3D models with Gaussian noise aug., stdev+entropy uncertainty. Blue/red dots denote inliers/outliers (Sec.\ \ref{sec:correlation}).} \label{fig_hipo2_3d}
\end{figure}

\noindent \textbf{\underline{3D models:}} Here, we used Gaussian noise augmentation, and estimated uncertainty with stdev and entropy. For \underline{anterior hippocampus} (not shown for brevity), the mean Dice was between 0.83-0.84. Inlier-outlier Dice difference was 0.05-0.08 with confidence, and 0.05-0.07 with VVC. For \underline{posterior hippocampus} (Fig.~\ref{fig_hipo2_3d}), the mean Dice was between 0.79-0.81.  Inlier-outlier Dice difference was 0.07-0.15 with confidence, and 0.07-0.14 with VVC. The inlier-outlier differences were significant for each fold ($p<0.05$) for both C and VVC, in both 2D and 3D models.

\begin{figure}[t]
\includegraphics[width=8.5cm]{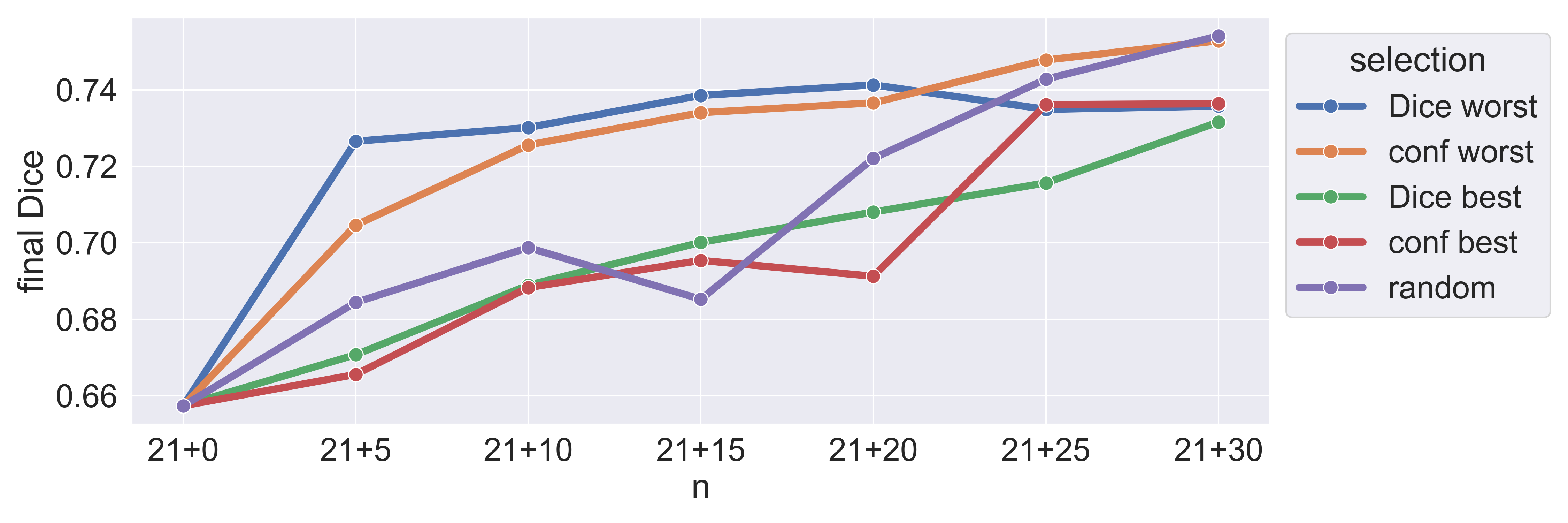}
\caption{Candidate selection strategies for manual segmentation. Selecting images with the worst confidence significantly outperforms random selection, and approaches the oracle strategy (selection by worst Dice).} \label{fig_conf_select}
\end{figure}

\noindent \textbf{\underline{Selection by confidence:}} 
Fig.~\ref{fig_conf_select} shows the mean test performance for each selection strategy, where $N=0$ is the base model. Adding images with the worst Dice/confidence scores improves test performance faster than other strategies. The Dice scores of the ``worst confidence'' strategy were significantly better than the ``random'' strategy for the smaller N ($p<0.01$ for $N\in\{5,...,20\}$, and $p<0.03$ for $N=25$ in paired t-tests). The performances converge, as expected, as we approach the entire candidate pool size, with the random strategy performing better at $N=30$ ($p<0.03$).

\noindent \textbf{\underline{Discussion:}} 
We proposed a novel method to estimate a single confidence score for CNNs used in medical image segmentation, 
by examining how rapidly the segmentation volume decreases as we gradually allow less uncertainty.

Our confidence score $C$ is highly correlated with the segmentation Dice. We observe that the upper left quadrant of the plots in Figs.\ \ref{fig-2d} and \ref{fig_hipo2_3d} is typically empty, i.e., $C$ is rarely high for low-Dice predictions in our experiments. This suggests high sensitivity, which is useful for practical workflows.
$C$ values are usually in the range of 0.50-0.90, and most cases with low Dice  receive a $C$ below 0.70. The unsupervised outlier detection experiment confirms this finding.
For each ROI in Fig.\ \ref{fig-2d}, VVC fails to flag several low-Dice ($<0.6$) cases that our method successfully captures (red arrows). Compared to VVC, our method has more consistent performance for a given task: e.g., for the anterior hippocampus (Fig.\ \ref{fig-2d}, middle row), the VVC correlations fluctuate from -0.247 to -0.668, whereas our correlations are stable in the [0.757, 0.785] range. Our correlations and inlier-outlier differences substantially surpass VVC counterparts in most experiments.

In our experiments we used test-time flips and Gaussian noise. These or other suitable augmentations may be used for new tasks. 
More extensive evaluation remains as future work.

Our proposed confidence score $C$  shows a strong correlation with Dice in our extensive experiments, reliably indicating segmentation quality.
Our method's compatibility with existing segmentation pipelines and its applicability without requiring specialized architectures make it a versatile tool. This technique can help in many scenarios such as identifying poor-quality inputs or for selecting images for manual segmentation. Overall, our approach contributes a valuable layer of uncertainty estimation and quality control, enhancing the potential for increased clinical adoption of CNNs. Application in clinic remains as future work. Our code is available at \url{github.com/MedICL-VU/segmentation-confidence}.

\noindent \textbf{\underline{Acknowledgments:}} 
This work was supported, in part, by NIH R01HD109739, U01-HD087180, R03-HD-069742, and Penn Presbyterian George L. and Emily McMichael Harrison Fund for Research in Obstetrics and Gynecology.

\bibliographystyle{IEEEbib}
\bibliography{mybibliography}

\begin{thebibliography}{10}

\bibitem{long2015fully}
J~Long, E~Shelhamer, and T~Darrell,
\newblock ``Fully convolutional networks for semantic segmentation,''
\newblock in {\em IEEE CVPR}, 2015, pp. 3431--3440.

\bibitem{ronneberger2015u}
O~Ronneberger, P~Fischer, and T~Brox,
\newblock ``U-net: Convolutional networks for biomedical image segmentation,''
\newblock in {\em MICCAI}. Springer, 2015, pp. 234--241.

\bibitem{he2016deep}
K~He, X~Zhang, S~Ren, and J~Sun,
\newblock ``Deep residual learning for image recognition,''
\newblock {\em CVPR}, pp. 770--778, 2016.

\bibitem{huang2017densely}
G~Huang, Z~Liu, L~Van Der~Maaten, and KQ~Weinberger,
\newblock ``Densely connected convolutional networks,''
\newblock in {\em CVPR}, 2017, pp. 4700--4708.

\bibitem{chollet2017xception}
F~Chollet,
\newblock ``Xception: Deep learning with depthwise separable convolutions,''
\newblock {\em CVPR}, pp. 1251--1258, 2017.

\bibitem{oktay2018attention}
O~Oktay, J~Schlemper, LL~Folgoc, M~Lee, M~Heinrich, K~Misawa, K~Mori,
  S~McDonagh, NY~Hammerla, B~Kainz, et~al.,
\newblock ``Attention u-net: Learning where to look for the pancreas,''
\newblock {\em arXiv:1804.03999}, 2018.

\bibitem{isensee2021nnu}
F~Isensee, PF~Jaeger, SAA Kohl, J~Petersen, and KH~Maier-Hein,
\newblock ``{nnU-Net: a self-configuring method for deep learning-based
  biomedical image segmentation},''
\newblock {\em Nature methods}, vol. 18, pp. 203--211, 2021.

\bibitem{hasani2022trustworthy}
N~Hasani, MA~Morris, A~Rhamim, RM~Summers, E~Jones, E~Siegel, and B~Saboury,
\newblock ``{Trustworthy AI in medical imaging},''
\newblock {\em PET clinics}, vol. 17, no. 1, 2022.

\bibitem{chen2022explainable}
H~Chen, C~Gomez, CM~Huang, and M~Unberath,
\newblock ``Explainable medical imaging ai needs human-centered design:
  guidelines and evidence from a systematic review,''
\newblock {\em NPJ digital medicine}, vol. 5, no. 1, pp. 156, 2022.

\bibitem{jin2023guidelines}
W~Jin, X~Li, M~Fatehi, and G~Hamarneh,
\newblock ``Guidelines and evaluation of clinical explainable ai in medical
  image analysis,''
\newblock {\em Medical image analysis}, 2023.

\bibitem{czolbe2021segmentation}
S~Czolbe, K~Arnavaz, O~Krause, and A~Feragen,
\newblock ``Is segmentation uncertainty useful?,''
\newblock {\em IPMI}, 2021.

\bibitem{gal2016dropout}
Y~Gal and Z~Ghahramani,
\newblock ``Dropout as a bayesian approximation: Representing model uncertainty
  in deep learning,''
\newblock in {\em ICML}. PMLR, 2016, pp. 1050--1059.

\bibitem{ghoshal2021estimating}
B~Ghoshal, A~Tucker, B~Sanghera, and W~Lup~Wong,
\newblock ``Estimating uncertainty in deep learning for reporting confidence to
  clinicians in medical image segmentation and diseases detection,''
\newblock {\em Computational Intelligence}, vol. 37, no. 2, pp. 701--734, 2021.

\bibitem{teye2018bayesian}
M~Teye, H~Azizpour, and K~Smith,
\newblock ``Bayesian uncertainty estimation for batch normalized deep
  networks,''
\newblock in {\em ICML}. PMLR, 2018, pp. 4907--4916.

\bibitem{wang2019aleatoric}
G~Wang, W~Li, M~Aertsen, J~Deprest, S~Ourselin, and T~Vercauteren,
\newblock ``{Aleatoric uncertainty estimation with test-time augmentation for
  medical image segmentation with CNNs},''
\newblock {\em Neurocomputing}, 2019.

\bibitem{lakshminarayanan2017simple}
B~Lakshminarayanan, A~Pritzel, and C~Blundell,
\newblock ``Simple and scalable predictive uncertainty estimation using deep
  ensembles,''
\newblock {\em NeurIPS}, vol. 30, 2017.

\bibitem{karimi2019accurate}
D~Karimi, Q~Zeng, P~Mathur, A~Avinash, S~Mahdavi, I~Spadinger, P~Abolmaesumi,
  and SE~Salcudean,
\newblock ``Accurate and robust deep learning-based segmentation of the
  prostate clinical target volume in ultrasound images,''
\newblock {\em Medical image analysis}, vol. 57, pp. 186--196, 2019.

\bibitem{kohl2018probabilistic}
S~Kohl, B~Romera-Paredes, C~Meyer, J~De~Fauw, JR~Ledsam, K~Maier-Hein,
  SM~Eslami, DJ~Rezende, and O~Ronneberger,
\newblock ``A probabilistic u-net for segmentation of ambiguous images,''
\newblock {\em NeurIPS}, vol. 31, 2018.

\bibitem{yao2023false}
X~Yao, H~Liu, D~Hu, D~Lu, A~Lou, H~Li, R~Deng, G~Arenas, B~Oguz, N~Schwartz,
  et~al.,
\newblock ``False negative/positive control for sam on noisy medical images,''
\newblock {\em arXiv:2308.10382}, 2023.

\bibitem{akrami2021quantile}
H~Akrami, A~Joshi, S~Aydore, and R~Leahy,
\newblock ``{Quantile regression for uncertainty estimation in VAEs with
  applications to brain lesion detection},''
\newblock {\em IPMI}, 2021.

\bibitem{mehrtash2020confidence}
A~Mehrtash, WM~Wells, CM~Tempany, P~Abolmaesumi, and T~Kapur,
\newblock ``Confidence calibration and predictive uncertainty estimation for
  deep medical image segmentation,''
\newblock {\em IEEE TMI}, vol. 39, no. 12, pp. 3868--3878, 2020.

\bibitem{liu2023colossal}
H~Liu, H~Li, X~Yao, Y~Fan, D~Hu, BM~Dawant, V~Nath, Z~Xu, and I~Oguz,
\newblock ``Colossal: A benchmark for cold-start active learning for 3d medical
  image segmentation,''
\newblock in {\em MICCAI}. Springer, 2023, pp. 25--34.

\bibitem{nath2020diminishing}
V~Nath, D~Yang, BA~Landman, D~Xu, and HR~Roth,
\newblock ``Diminishing uncertainty within the training pool: Active learning
  for medical image segmentation,''
\newblock {\em IEEE TMI}, vol. 40, no. 10, pp. 2534--2547, 2020.

\bibitem{schwartz2022fully}
N~Schwartz, I~Oguz, J~Wang, A~Pouch, N~Yushkevich, S~Parameshwaran, J~Gee,
  P~Yushkevich, and B~Oguz,
\newblock ``Fully automated placental volume quantification from 3d ultrasound
  for prediction of small-for-gestational-age infants,''
\newblock {\em JUM}, vol. 41, no. 6, pp. 1509--1524, 2022.

\bibitem{simpson2019large}
A~Simpson et~al.,
\newblock ``A large annotated medical image dataset for the development and
  evaluation of segmentation algorithms,''
\newblock {\em arXiv:1902.09063}, 2019.

\end{thebibliography}

\end{document}